\documentclass[acmtog]{acmart}
\acmSubmissionID{766}

\copyrightyear{2026}
\acmYear{2026}
\setcopyright{cc}
\setcctype{by}
\acmConference[SIGGRAPH Conference Papers '26]{Special Interest Group on Computer Graphics and Interactive Techniques Conference Conference Papers}{July 19--23, 2026}{Los Angeles, CA, USA}
\acmBooktitle{Special Interest Group on Computer Graphics and Interactive Techniques Conference Conference Papers (SIGGRAPH Conference Papers '26), July 19--23, 2026, Los Angeles, CA, USA}
\acmDOI{10.1145/3799902.3811134}
\acmISBN{979-8-4007-2554-8/2026/07}

\usepackage{booktabs} 

\usepackage[ruled]{algorithm2e} 
\usepackage{xcolor}
\usepackage{multirow}
\usepackage{enumitem}
\usepackage{float}
\usepackage{listings}

\definecolor{keyblue}{rgb}{0.0, 0.0, 0.4}
\definecolor{strgray}{rgb}{0.2, 0.2, 0.2}

\lstdefinelanguage{json}{
    basicstyle=\normalfont\ttfamily,
    string=[b]",
    stringstyle=\color{strgray},
    comment=[l]{//},
    commentstyle=\color{gray}\itshape,
    morekeywords={true,false,null},
    keywordstyle=\color{keyblue}\bfseries,
    sensitive=true
}

\lstdefinestyle{academicstyle}{
    language=json,
    basicstyle=\ttfamily\scriptsize,
    columns=flexible,
    keepspaces=true,
    breaklines=true,
    breakatwhitespace=false,
    frame=tb,
    rulecolor=\color{black},
    backgroundcolor=,
    numbers=none,
    showstringspaces=false,
    captionpos=b,
    aboveskip=0.8em,
    belowskip=0.8em,
    xleftmargin=0.5em,
    xrightmargin=0.5em,
    literate={\ \ }{{\ }}1
}

\SetAlFnt{\small}
\SetAlCapFnt{\small}
\SetAlCapNameFnt{\small}
\SetAlCapHSkip{0pt}

\def\sysname{AGILE}

\definecolor{RevisionBlue}{RGB}{0,70,200}
\newcommand{\revision}[1]{#1}

\definecolor{TableDarkGreen}{RGB}{182,215,168}
\definecolor{TableLightGreen}{RGB}{207,234,215}
\definecolor{TableYellow}{RGB}{255,242,204}
\definecolor{TableRed}{RGB}{244,204,204}

\begin{document}
\title{\sysname: Hand-Object Interaction Reconstruction from Video via Agentic Generation}

\author{Jin-Chuan Shi}
\authornote{Equal contribution.}
\affiliation{%
  \institution{State Key Lab of CAD \& CG, Zhejiang University}
  \city{Hangzhou}
  \country{China}}

\author{Binhong Ye}
\authornotemark[1]
\affiliation{%
  \institution{State Key Lab of CAD \& CG, Zhejiang University}
  \city{Hangzhou}
  \country{China}}

\author{Tao Liu}
\affiliation{%
  \institution{State Key Lab of CAD \& CG, Zhejiang University}
  \city{Hangzhou}
  \country{China}}

\author{Junzhe He}
\affiliation{%
  \institution{State Key Lab of CAD \& CG, Zhejiang University}
  \city{Hangzhou}
  \country{China}}

\author{Yangjinhui Xu}
\affiliation{%
  \institution{State Key Lab of CAD \& CG, Zhejiang University}
  \city{Hangzhou}
  \country{China}}

\author{Xiaoyang Liu}
\affiliation{%
  \institution{State Key Lab of CAD \& CG, Zhejiang University}
  \city{Hangzhou}
  \country{China}}

\author{Zeju Li}
\affiliation{%
  \institution{State Key Lab of CAD \& CG, Zhejiang University}
  \city{Hangzhou}
  \country{China}}

\author{Hao Chen}
\authornote{Corresponding author.}
\affiliation{%
  \institution{State Key Lab of CAD \& CG, Zhejiang University}
  \city{Hangzhou}
  \country{China}}

\author{Chunhua Shen}
\affiliation{%
  \institution{State Key Lab of CAD \& CG, Zhejiang University}
  \city{Hangzhou}
  \country{China}}
\affiliation{%
  \institution{Zhejiang University of Technology}
  \city{Hangzhou}
  \country{China}}

\renewcommand\shortauthors{Shi et al.}

\begin{abstract}
Reconstructing dynamic hand-object interactions from monocular videos is critical for dexterous manipulation data collection and creating realistic digital twins for robotics and VR. However, current methods face two prohibitive barriers: (1) reliance on neural rendering often yields fragmented, non-simulation-ready geometries under heavy occlusion, and (2) dependence on brittle Structure-from-Motion (SfM) initialization leads to frequent failures on in-the-wild footage. To overcome these limitations, we introduce \sysname{}, a robust framework that shifts the paradigm from \textit{reconstruction} to \textit{agentic generation} for interaction learning. First, we employ an agentic pipeline where a Vision-Language Model (VLM) guides a generative model to synthesize a complete, watertight object mesh with high-fidelity texture, independent of video occlusions. Second, bypassing fragile SfM entirely, we propose a robust \textit{anchor-and-track} strategy. We initialize the object pose at a single interaction onset frame using a foundation model and propagate it temporally by leveraging the strong visual similarity between our generated asset and video observations. Finally, a contact-aware optimization integrates semantic, geometric, and interaction stability constraints to enforce physical plausibility. Extensive experiments on \revision{HO3D, DexYCB, ARCTIC, and in-the-wild videos} reveal that \sysname{} outperforms baselines in global geometric accuracy while demonstrating exceptional robustness on challenging sequences where prior arts frequently collapse. By prioritizing physical validity, our method produces simulation-ready assets validated via real-to-sim retargeting for robotic applications.
Project page: \href{https://agile-hoi.github.io/}{agile-hoi.github.io}.
\end{abstract}

%
%


%
%

\begin{teaserfigure}
  \centering
  \includegraphics[width=\textwidth]{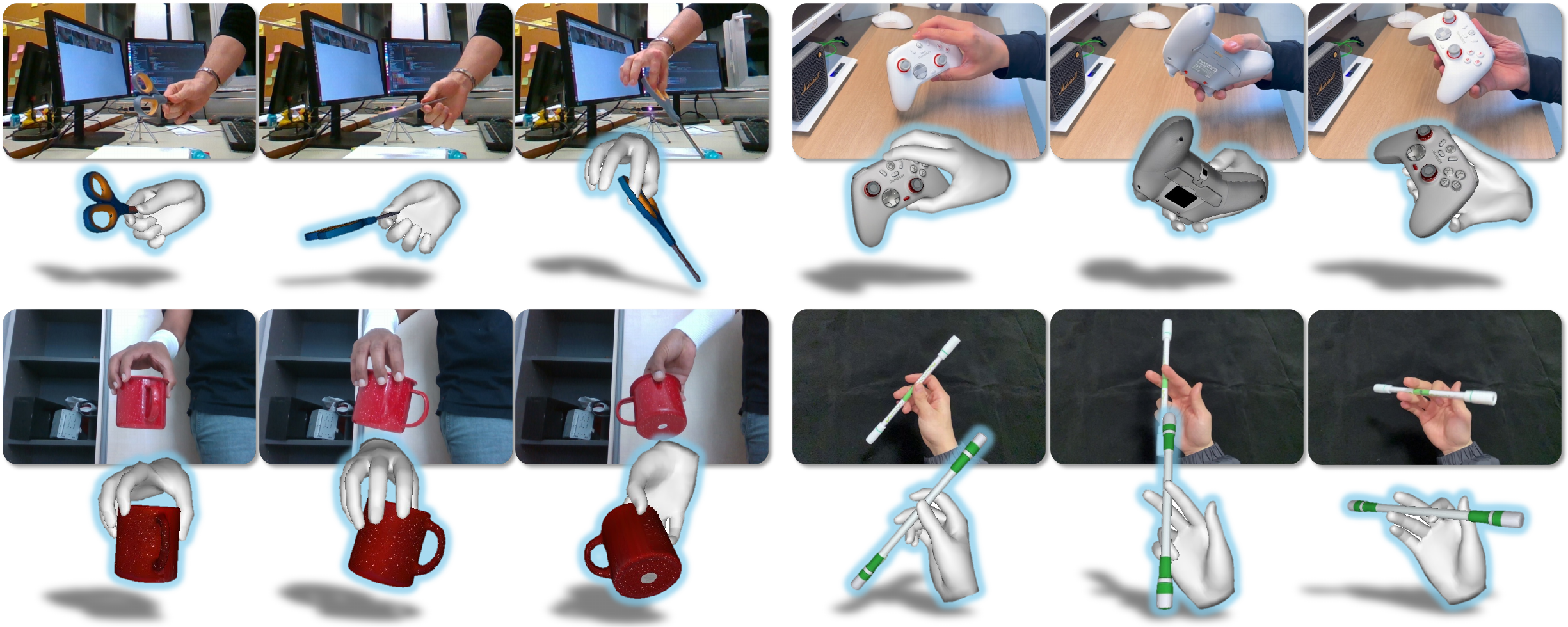}
  \caption{\textbf{High-Fidelity Hand-Object Reconstruction from Video.}
  We present \sysname{}, a framework that reconstructs simulation-ready interaction sequences from monocular video.
  By leveraging agentic generative priors, \sysname{} robustly recovers watertight geometry, realistic textures, and precise 6D poses for diverse objects, ranging from thin structures (scissors, pen) to complex topologies (game controller), even under severe hand occlusion and rapid manipulation.
  }
  \label{fig:teaser}
\end{teaserfigure}

\maketitle

\section{Introduction}
\label{sec:intro}

Reconstructing dynamic hand-object interactions (HOI) from monocular video is a pivotal challenge in computer vision and graphics. The ultimate aspiration is to create high-fidelity digital twins of these interactions, assets that are not only visually coherent but also geometrically explicit and physically plausible for downstream tasks such as robotic imitation learning and physics-based simulation. While the internet offers a vast repository of in-the-wild interaction videos, converting this data into simulation-ready 3D assets remains an unresolved challenge. We identify two fundamental barriers in existing pipelines that hinder scalability and reliability.

The first barrier lies in the inherent limitations of \textit{reconstruction-based geometry}. State-of-the-art methods typically rely on neural rendering techniques, such as NeRF~\cite{mildenhall2021nerf} or 3D Gaussian Splatting~\cite{kerbl20233d}, to optimize scene representations~\cite{fan2024hold, on2025bigs}. However, these methods fundamentally depend on multi-view consistency. In HOI scenarios, severe hand occlusion frequently violates this premise, leading to fragmented, noisy, or non-watertight geometries. Such representations, while visually passable from trained viewpoints, are ill-suited for physics engines that require clean, explicit topology.

The second, and more critical, bottleneck is the \textit{brittleness of pose initialization}. Most prevailing approaches~\cite{fan2024hold, wang2025magichoi} are anchored on Structure-from-Motion (SfM) pipelines like COLMAP~\cite{schoenberger2016sfm} to estimate initial camera and object poses. SfM is notoriously fragile in dynamic scenes featuring textureless objects, rapid motions, or significant occlusions. When SfM fails to register frames, a common occurrence in real-world footage, the entire pipeline collapses. This reliance on classic geometric matching prevents current methods from achieving reliable performance on complex benchmarks~\cite{chao2021dexycb}.

Recent generative advancements offer potential but lack robustness for high-fidelity HOI. MagicHOI~\cite{wang2025magichoi} leverages diffusion priors but produces over-smoothed meshes and retains a dependency on brittle SfM. Single-view methods like SAM3D~\cite{sam3dteam2025sam3d3dfyimages} suffer from occlusion-induced information loss, yielding coarse geometries and poor textures that destabilize subsequent pose optimization. To overcome these barriers, it is essential to aggregate multi-view cues from the video. However, since standard 3D generators require canonical cameras unavailable in dynamic footage, we must resort to 2D generative priors to synthesize missing perspectives. Yet, the stochastic nature of 2D diffusion models frequently introduces hallucinations inconsistent with the actual object in the video. Consequently, reliably harnessing these priors requires an intelligent mechanism capable of identifying informative frames and strictly filtering generated views for consistency—a capability missing in current end-to-end pipelines.

To bridge this gap, we introduce \sysname{}, a robust framework that shifts the paradigm from \textit{reconstruction} to \textit{agentic generation} for interaction learning. First, we employ an agentic pipeline where a Vision-Language Model (VLM) acts as an intelligent supervisor. The VLM selects informative keyframes to guide multi-view synthesis and rigorously filters the outputs via rejection sampling to ensure consistency. Crucially, the pipeline incorporates a \textit{texture refinement} phase that recovers high-frequency details from the video observations. This yields a watertight mesh with hyper-realistic appearance—a prerequisite that enables our subsequent foundation models to reliably initialize pose and optimize semantic alignment.

Leveraging this high-quality asset, we propose a robust \textit{anchor-and-track} optimization strategy that eliminates the need for brittle SfM. We utilize a foundation model to initialize the object pose only at a single interaction onset frame. For the remainder of the video, we perform online optimization by propagating the pose temporally. By capitalizing on the strong visual similarity between our generated textured mesh and the video observations, we drive this alignment using semantic feature loss and mask constraints, while simultaneously enforcing physical contact consistency. This approach proves significantly more stable than relying on noisy per-frame predictions or fragile SfM.

Comprehensive evaluations across benchmarks (HO3D, DexYCB) and diverse in-the-wild videos demonstrate the transformative impact of our approach. \sysname{} achieves state-of-the-art performance across all datasets, establishing a new standard for both accuracy and robustness. While prior methods suffer from high failure rates (up to 75\%) on complex sequences, \sysname{} maintains a 100\% success rate. Our analysis reveals a critical trade-off: while some prior methods optimize for local proximity—often allowing physical interpenetration—our approach prioritizes global geometric accuracy and non-penetration. Consequently, \sysname{} delivers digital twins that are validated via a real-to-sim retargeting pipeline, successfully driving dynamic interactions in a physics simulator.

\revision{Our core contribution is a paradigm shift from reconstruction to \textit{agentic generation} for HOI, realized through the following designs.} First, \revision{and central to our novelty,} we propose the first agentic HOI pipeline that integrates VLM-guided quality assessment with generative models, enabling the production of high-fidelity, watertight meshes independent of video occlusions. Second, we introduce a robust anchor-and-track optimization strategy that eliminates the dependency on brittle SfM by anchoring pose initialization at a single contact frame and propagating it via semantic and geometric alignment. \revision{Third, extensive experiments across single-hand benchmarks, bimanual interactions, and in-the-wild sequences, together with real-to-sim retargeting on a dexterous robotic hand, validate that \sysname{} achieves state-of-the-art geometric accuracy, exceptional robustness, and produces simulation-ready digital twins.}
\section{Related Work}
\label{sec:related_work}

\paragraph{3D Hand Reconstruction.}
Monocular 3D hand reconstruction has evolved from fitting parametric models like MANO~\cite{romero2022embodied} to images~\cite{boukhayma20193d, pavlakos2024reconstructing, zhang2019end, zhou2020monocular}, towards recovering dynamic motion from video. While recent temporal approaches~\cite{yu2025dyn, zhang2025hawor} regress 4D trajectories, they often sacrifice the per-frame geometric precision essential for contact analysis. In our framework, we prioritize geometric fidelity to ensure accurate interaction modeling. Consequently, we leverage state-of-the-art estimators like WiLoR~\cite{potamias2025wilor} to provide robust, high-fidelity initial hand meshes.

\paragraph{Generative 3D Object Reconstruction.}
The field is shifting from multi-view optimization (e.g., NeRF~\cite{mildenhall2021nerf}, Gaussian Splatting~\cite{kerbl20233d})—which requires dense, unoccluded coverage rarely found in interaction videos—to generative inference. Large Reconstruction Models (LRMs)~\cite{hong2023lrm, tang2024lgm, xu2024instantmesh, hunyuan3d22025tencent} and diffusion-based methods~\cite{liu2023zero, long2024wonder3d} now generate assets from sparse inputs. However, single-view methods like SAM3D~\cite{sam3dteam2025sam3d3dfyimages}, while capable of end-to-end inference, inherently lack the capacity to aggregate multi-view cues from video. This limitation prevents them from resolving severe occlusions or maintaining texture consistency across views. We address this by proposing an agentic framework that intelligently fuses video evidence with generative priors, synthesizing watertight, textured assets that strictly align with video observations.

\paragraph{Hand-Object Interaction (HOI) Reconstruction.}
Reconstructing dynamic HOI is complicated by severe mutual occlusion. Early template-based~\cite{tekin2019h+, corona2020ganhand, yang2021cpf, fan2024benchmarks} or depth-dependent methods limit in-the-wild applicability. 
\textit{Reconstruction-based methods} like HOLD~\cite{fan2024hold} and BIGS~\cite{on2025bigs} optimize implicit representations but yield fragmented, non-physical geometries under occlusion. Furthermore, their reliance on brittle Structure-from-Motion (SfM) initialization (e.g., COLMAP~\cite{schoenberger2016sfm}) leads to frequent failures on textureless or moving objects.
\textit{Generative approaches} attempt to hallucinate missing regions; MagicHOI~\cite{wang2025magichoi} integrates diffusion priors with NeRF but relies on Score Distillation Sampling (SDS), often producing over-smoothed meshes lacking simulation-ready detail. Crucially, it retains the fragility of SfM initialization. In contrast, \sysname{} eliminates the SfM bottleneck via a robust anchor-and-track strategy, delivering high-fidelity, simulation-ready digital twins.
\section{Method}
\label{sec:method}

\begin{figure*}[t] 
\centering
\includegraphics[width=\linewidth]{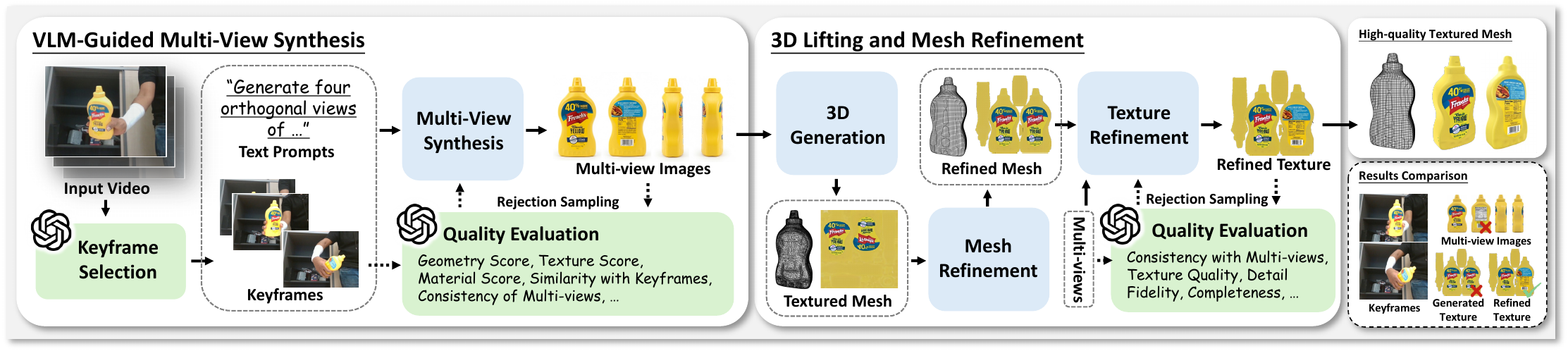}
    \caption{
    \textbf{Pipeline for Agentic Textured Object Generation.} 
    A VLM agent first selects informative keyframes from the input video to guide multi-view synthesis. 
    To ensure consistency, a VLM-based critic filters the generated views via rejection sampling. 
    The validated images are then lifted to 3D, followed by automated topology optimization and texture refinement. 
    As highlighted in the bottom-right comparison, this refinement step significantly enhances texture fidelity against the evaluated multi-views, yielding a high-quality, simulation-ready asset.
    } 
    \label{fig:generation_pipe}
\end{figure*}

\begin{figure*}[t]
    \centering
    \includegraphics[width=\linewidth]{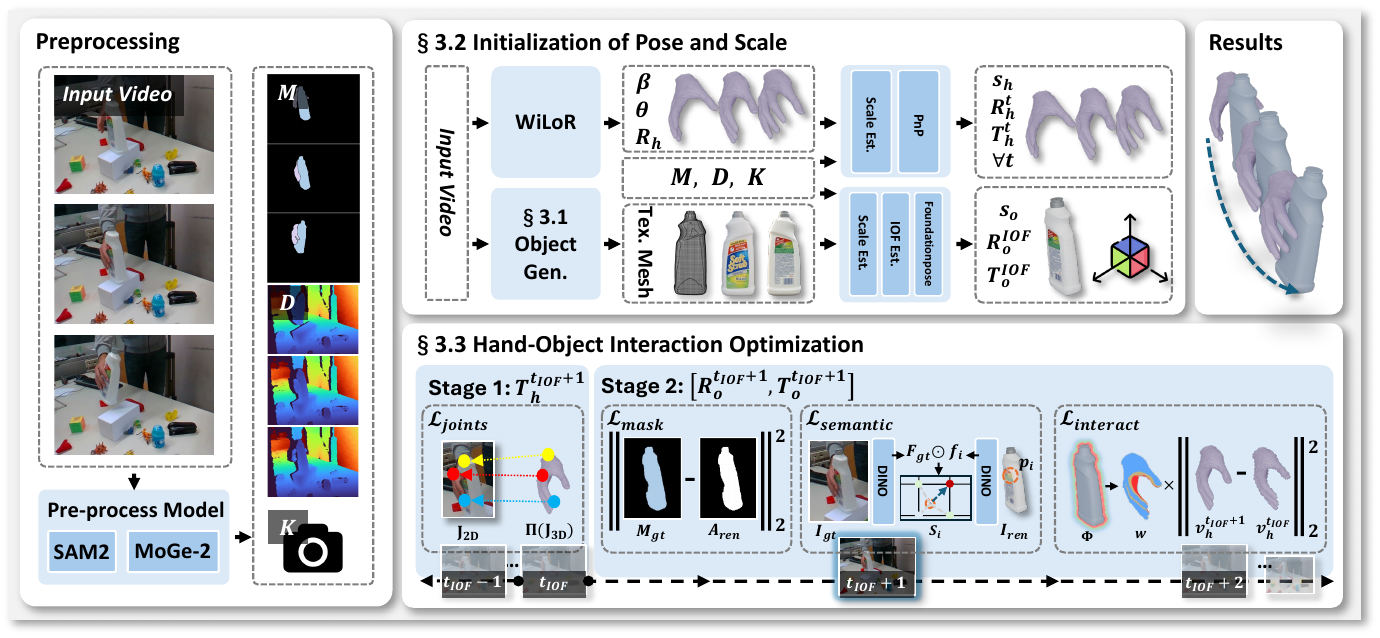}
    \caption{
    \textbf{Pipeline of \sysname{}.} 
    Our framework processes the input video in three phases:
    (1) \textbf{Agentic Generation (§3.1):} A VLM-guided loop extracts keyframes and supervises the synthesis of a watertight, textured object mesh $\mathcal{M}_o$, utilizing rejection sampling to ensure visual fidelity.
    (2) \textbf{SfM-Free Initialization (§3.2):} We decouple metric scale and pose. The hand is initialized via WiLoR, while the object pose is anchored at the Interaction Onset Frame (IOF) using a foundation model.
    (3) \textbf{Contact-Aware Optimization (§3.3):} A bi-directional tracking process refines the trajectories. We stabilize the hand via geometric alignment and track the object using semantic ($\mathcal{L}_{dino}$) and interaction constraints ($\mathcal{L}_{interact}$) to ensure physical plausibility.
    }
    \label{fig:pipe}
\end{figure*}

Given a monocular video of a hand-object interaction captured by a fixed camera, our goal is to reconstruct the 4D trajectory of both the hand and the object, yielding high-fidelity, simulation-ready 3D assets, as illustrated in Figure~\ref{fig:pipe}.
Unlike previous methods~\cite{fan2024hold, on2025bigs} that rely on neural rendering and fragile Structure-from-Motion (SfM) initialization, we propose \sysname{}, a generative framework that shifts the paradigm from \textit{reconstruction} to \textit{agentic generation} for hand-object interaction reconstruction.

\subsection{Agentic Textured Object Generation}
\label{sec:object_generation}

A primary challenge in reconstructing hand-object interactions from monocular video is the severe occlusion of the object by the user's hand. While neural rendering~\cite{mildenhall2021nerf, kerbl20233d} excels at view synthesis, it struggles to recover complete geometry from such partially visible data. Similarly, single-view generative methods like SAM3D~\cite{sam3dteam2025sam3d3dfyimages} inherently suffer from information loss, yielding coarse geometries and low-fidelity textures that destabilize subsequent pose optimization. To overcome these limitations, it is essential to aggregate multi-view cues from the video using 2D generative priors to synthesize missing perspectives. However, the stochastic nature of diffusion models frequently introduces hallucinations that deviate from the actual object in the video. Reliably harnessing these priors therefore requires an intelligent mechanism capable of identifying the most informative reference frames and strictly filtering generated views for consistency—a capability missing in standard pipelines. To address this, we propose an agentic framework where a VLM supervisor acts as this intelligent critic, bridging the gap between noisy generative priors and rigorous video evidence.

\paragraph{VLM-Guided Multi-View Synthesis.}
As illustrated in Figure~\ref{fig:generation_pipe}, our pipeline operates through a cascade of generative models steered by a VLM agent. Instead of relying on a single frame, the VLM first selects $N$ informative keyframes (typically 1--4) from the input video to maximize viewpoint coverage. These frames prompt an image generation model to synthesize orthogonal views of the object. 
To ensure these hallucinations faithfully reflect the video content, we introduce a VLM-based critic. The VLM evaluates the consistency between generated views and original video frames, scoring them on geometry, texture, and material correspondence. Through a rejection sampling process, candidates falling below a strict consistency threshold are discarded and re-generated. This ensures that the input for 3D lifting is not only multi-view consistent but also faithful to the real-world observation. Implementation details are provided in the Supplementary Material.

\paragraph{3D Lifting and Mesh Refinement.}
The validated multi-view images are processed by a feed-forward 3D generation model~\cite{hunyuan3d22025tencent} to produce an initial mesh. While geometrically plausible, raw outputs often suffer from irregular topology and blurry textures. To address this, we first apply automated retopology and UV unwrapping to create a clean, lightweight mesh optimized for physics simulation. 
Subsequently, we perform an \textit{agentic texture refinement} step. The initial texture is enhanced using an image-to-image editing model conditioned on the evaluated high-resolution multi-view images to recover details.Crucially, as shown in Figure~\ref{fig:generation_pipe}, this process is also supervised by the VLM agent. The VLM critic evaluates the refined texture against the multi-views via rejection sampling, ensuring strict visual fidelity and discarding any hallucinated artifacts.
This rigorous quality control is motivated by two downstream requirements: (1) Foundation models used for pose initialization (e.g., FoundationPose~\cite{wen2024foundationpose}) rely heavily on sharp textual features for robust matching; and (2) our subsequent optimization utilizes DINO features, which demand high visual fidelity to maintain accuracy. This refinement yields a simulation-ready asset with photo-realistic appearance and explicit topology.

Our agentic framework offers distinct advantages over traditional baselines. First, it enables fully automated asset acquisition from arbitrary videos, effectively bypassing the failure modes of photogrammetry under heavy occlusion. Second, by continually aggregating temporal cues via VLM supervision, we ensure exceptional visual consistency between the generated asset and the video. Finally, unlike implicit representations, our approach directly produces clean, watertight meshes with high-frequency textures, providing a robust initialization that is critical for the stability of subsequent interaction tracking and physics simulation.

\subsection{Initialization of Pose and Scale}
\label{sec:init_pose_scale}

To enable robust tracking without relying on brittle Structure-from-Motion pipelines, we establish a consistent metric initialization for both the hand and object. 
First, we extract essential scene information, including segmentation masks, camera intrinsics, and estimated metric depth. Second, we recover the metric geometry of the hand, comprising its 6D pose, global scale, and 2D joint projections. Finally, we estimate the object's global scale across the sequence and initialize its pose at the interaction onset frame.

\paragraph{Data Preprocessing.}
For each input video frame, we extract a set of key components. Specifically, we employ the monocular metric depth estimator MoGe-2~\cite{wang2025moge} to obtain the camera intrinsic matrix $\mathbf{K}$ and a per-frame metric depth map $\mathbf{D} \in \mathbb{R}^{H \times W}$. Simultaneously, we use SAM2~\cite{ravi2024sam} to acquire precise segmentation masks for both the hand ($\mathbf{M}_h$) and the object ($\mathbf{M}_o$). Given our static camera assumption, the camera coordinate system serves as the world frame, with extrinsic parameters fixed as the identity matrix. These preprocessed priors form the foundational inputs for all subsequent stages.

\paragraph{Metric Hand Initialization.}
We utilize the preprocessed data to initialize a physically plausible hand model. We first employ the off-the-shelf estimator WiLoR~\cite{potamias2025wilor} to predict the MANO~\cite{romero2022embodied} parameters for each frame, including shape $\boldsymbol{\beta}$, pose $\boldsymbol{\theta}$, global rotation $\mathbf{R}_h \in SO(3)$, and the corresponding 2D keypoint annotations $\mathbf{J}_{2D} \in \mathbb{R}^{21 \times 2}$. Since WiLoR provides reliable rotation estimates, we fix $\mathbf{R}_h$ and focus on recovering the missing metric scale and translation.
To determine the global scale $s_h \in \mathbb{R}^+$, we unproject the masked hand pixels into 3D point clouds using the metric depth map $\mathbf{D}$. We align the MANO mesh to these depth observations using a constrained Iterative Closest Point (ICP) algorithm~\cite{chetverikov2002trimmed}. In this step, we hold $\mathbf{R}_h$ fixed and optimize for $s_h$ and a temporary translation to minimize the model-to-scan distance. Finally, to ensure accurate image alignment, we fix $s_h$, $\mathbf{R}_h$, and the intrinsics $\mathbf{K}$, and solve for the per-frame translation $\mathbf{T}_h \in \mathbb{R}^3$ via a Perspective-n-Point (PnP) approach, utilizing the 3D model joints and the predicted 2D keypoints $\mathbf{J}_{2D}$.

\paragraph{Object Pose and Scale Estimation.}
Once the canonical object mesh is created (Sec.~\ref{sec:object_generation}), we align it to the video observations using a scale-first strategy. 
We first estimate the object's global metric scale $s_o$ by applying the same constrained ICP algorithm used for the hand, aligning the generated mesh with the unprojected object point clouds across all available frames.
Next, we identify the \textit{interaction onset frame} (IOF), defined as the moment the object mask $\mathbf{M}_o$ exhibits significant displacement.
Finally, to determine the initial object pose $[\mathbf{R}_o^{IOF}, \mathbf{T}_o^{IOF}]$, we apply FoundationPose~\cite{wen2024foundationpose} at the IOF. This estimator takes the RGB frame, the metric depth map, and the generated mesh pre-scaled by $s_o$ as input, ensuring the estimated pose is metrically consistent with the scene.

\subsection{Hand-Object Interaction Optimization}
\label{sec:joint_optimization}

To ensure pixel-level alignment and physical consistency, we employ a bi-directional online optimization strategy anchored at the \textit{interaction onset frame (IOF)} (identified in Sec.~\ref{sec:init_pose_scale}). 
Starting from the IOF, we propagate the optimization frame-by-frame towards both the start and end of the video sequence.
At the IOF, we jointly optimize the hand and object poses along with the object's anisotropic scale $\mathbf{s}_o$ to establish a reliable canonical geometry. For all subsequent frames, $\mathbf{s}_o$ is frozen, and we perform the following two-step optimization per frame:

\paragraph{Step 1: Hand Translation Refinement.}
For the current frame, we first refine the hand trajectory. Relying on the robust orientation estimates from the initialization stage, we fix the hand rotation $\mathbf{R}_h$ and scale $s_h$. We exclusively optimize the translation $\mathbf{T}_h$ to align the mesh with 2D observations. This step is driven solely by the \textit{joint reprojection loss} ($\mathcal{L}_{\text{joint}}$), which minimizes the Euclidean distance between projected 3D joints and 2D detections. This strictly geometric update ensures the hand serves as a stable anchor for the object.

\paragraph{Step 2: Interaction-Aware Object Tracking.}
With the refined hand pose fixed, we optimize the object's rigid pose $(\mathbf{R}_o, \mathbf{T}_o)$. To handle rapid motion or occlusion, the object pose is initialized using the result from the preceding processed frame.
During this stage, we minimize a composite objective function to ensure visual and physical fidelity:
\begin{equation}
    \mathcal{L}_{\text{obj}} = 
    \lambda_{\text{mask}} \mathcal{L}_{\text{mask}} + 
    \lambda_{\text{dino}} \mathcal{L}_{\text{dino}} +
    \lambda_{\text{interact}} \mathcal{L}_{\text{interact}}.
\end{equation}
Here, $\mathcal{L}_{\text{mask}}$ enforces silhouette alignment, and $\mathcal{L}_{\text{dino}}$ maintains semantic feature consistency using DINOv3~\cite{simeoni2025dinov3} to mitigate texture ambiguity. Crucially, $\mathcal{L}_{\text{interact}}$ penalizes interpenetration and encourages surface attraction, ensuring the object remains tightly locked to the hand in a physically plausible manner.

\paragraph{Joint Reprojection Loss ($\mathcal{L}_{\text{joint}}$).}
To spatially anchor the hand pose using reliable 2D cues, we introduce a joint reprojection loss. Let $\Pi(\cdot)$ denote the perspective projection function given the camera intrinsics. We constrain the projected positions of the 3D hand joints $\mathbf{J}_{3D}$ to match the preprocessed 2D keypoint annotations $\mathbf{J}_{2D} \in \mathbb{R}^{21 \times 2}$. The loss is defined as the mean squared error (MSE):
\begin{equation}
    \mathcal{L}_{\text{joint}} = \frac{1}{N_j} \sum_{i=1}^{N_j} \| \Pi(\mathbf{J}_{3D}^{(i)}) - \mathbf{J}_{2D}^{(i)} \|_2^2.
\end{equation}
This term ensures the estimated hand geometry remains consistent with the detected anatomical landmarks on the image plane.

\paragraph{Mask Alignment Loss ($\mathcal{L}_{\text{mask}}$).}
To ensure the reconstructed geometry aligns with the visual evidence, we employ a mask alignment loss. This term measures the discrepancy between the rendered alpha silhouette $\mathbf{A}_{\text{ren}}$ and the ground-truth segmentation mask $\mathbf{M}_{\text{gt}}$. We formulate this as the $L_2$ distance between the two maps:
\begin{equation}
    \mathcal{L}_{\text{mask}} = \| \mathbf{A}_{\text{ren}} - \mathbf{M}_{\text{gt}} \|_2^2.
\end{equation}
By minimizing this difference, the optimization drives the rendered mesh to tightly fit the observed pixel-level silhouette, effectively handling occlusion boundaries.

\paragraph{Semantic Feature Loss ($\mathcal{L}_{\text{dino}}$).}
To mitigate tracking ambiguities caused by partial occlusions or textureless regions, we leverage deep semantic features from a pre-trained Vision Transformer, specifically DINOv3~\cite{simeoni2025dinov3}. Unlike pixel-wise photometric losses, semantic features provide robust correspondences invariant to local lighting changes.

We establish explicit 3D-to-2D semantic correspondences through a sampling-based approach. Periodically, we sample a set of $N_p$ visible points $\mathcal{P}_{can} = \{\mathbf{p}_i\}_{i=1}^{N_p}$ on the canonical object surface. For each point $\mathbf{p}_i$, we compute a \textit{feature similarity map} $\mathbf{S}_i \in \mathbb{R}^{H_f \times W_f}$ by correlating its feature vector (extracted from the rendered view) against the dense feature map of the target ground-truth image $\mathbf{F}_{gt}$. This similarity map $\mathbf{S}_i$ encodes the likelihood of the point $\mathbf{p}_i$ corresponding to each spatial location in the target image.

During optimization, we transform these canonical points into the current camera frame using the estimated object pose $(\mathbf{R}_o, \mathbf{T}_o)$ and scale $\mathbf{s}_o$, and project them to obtain 2D coordinates $\mathbf{u}_i = \Pi(\mathbf{R}_o (\mathbf{s}_o \odot \mathbf{p}_i) + \mathbf{T}_o)$.
The loss is formulated to maximize the semantic similarity at the projected locations, weighted by an occlusion mask to ignore unreliable regions:
\begin{equation}
    \mathcal{L}_{\text{dino}} = - \frac{1}{|\mathcal{V}|} \sum_{i \in \mathcal{V}} \mathcal{M}_{occ}(\mathbf{u}_i) \cdot \text{sample}(\mathbf{S}_i, \mathbf{u}_i),
\end{equation}
where $\mathcal{V}$ denotes the set of currently visible sampled points, $\mathcal{M}_{occ}$ is the binary occlusion mask (0 for occluded regions), and $\text{sample}(\cdot)$ represents bilinear interpolation of the similarity map at coordinates $\mathbf{u}_i$. By maximizing this similarity, the optimization drives the object pose to align its semantic parts with the corresponding regions in the video frame.

\paragraph{Interaction Stability Loss ($\mathcal{L}_{\text{interact}}$).}
Given the severe occlusion of the object during manipulation, strictly geometric cues from the image are often insufficient. To resolve this, we impose a physical prior that the relative spatial configuration between the hand and the object remains stable during a grasp\revision{, drawing on similar contact stability assumptions explored in prior works~\cite{fan2024hold, wang2025magichoi}}. This enforces the assumption that the hand and object move approximately as a rigid aggregate during interaction.

Specifically, we formulate a interaction stability loss in the object's canonical coordinate system. Let $\mathbf{V}_h^{(t)} \in \mathbb{R}^{N \times 3}$ denote the hand vertices in the camera frame at time $t$, and let $(\mathbf{R}_o^{(t)}, \mathbf{T}_o^{(t)})$ be the estimated object pose. We map the current hand vertices into the current object's local frame via the inverse transformation:
\begin{equation}
    \tilde{\mathbf{v}}_{h,i}^{(t)} = (\mathbf{R}_o^{(t)})^\top (\mathbf{v}_{h,i}^{(t)} - \mathbf{T}_o^{(t)}),
\end{equation}
where $\mathbf{v}_{h,i}^{(t)}$ is the $i$-th vertex of the hand. 
We then minimize the displacement of these local coordinates relative to their positions in the previous tracked frame $(t-1)$. 

Crucially, to ensure the constraint applies only to the effective grasping regions, we compute a vertex-wise weight map $\mathbf{w} \in \mathbb{R}^N$ based on the object's Signed Distance Function (SDF), denoted as $\Phi(\cdot)$. We calculate the distance of the previous hand vertices to the object surface: $d_i = \max(0, \Phi(\tilde{\mathbf{v}}_{h,i}^{(t-1)}))$, and derive the weights via a soft gating function $w_i = 1 - \tanh(\sigma \cdot d_i)$, where $\sigma$ is a scaling factor controlling the sensitivity of the fall-off. This formulation assigns high weights ($w_i \approx 1$) to vertices in close proximity to the object surface while suppressing the influence of non-interacting fingers. The final loss is defined as:
\begin{equation}
    \mathcal{L}_{\text{interact}} = \frac{1}{N} \sum_{i=1}^{N} w_i \cdot \| \tilde{\mathbf{v}}_{h,i}^{(t)} - \tilde{\mathbf{v}}_{h,i}^{(t-1)} \|_2.
\end{equation}
Minimizing this term effectively ``locks'' the object to the grasping hand parts, preventing unnatural sliding or jitter when visual features are ambiguous.

In summary, this optimization framework prioritizes reconstruction fidelity and physical consistency. 
By directly optimizing explicit 3D meshes, our method allows for the seamless integration of 2D visual supervision with 3D interaction constraints, which is often difficult for implicit representations. 
This design effectively resolves tracking ambiguities caused by severe hand occlusions. 
Furthermore, our decoupled, interaction-aware strategy ensures that the reconstructed motion is not only pixel-aligned but also physically grounded, preserving the high-quality geometry and texture assets generated in the previous stage.

\section{Experiments}
\label{sec:experiments}

\paragraph{Datasets.}
We evaluate \sysname{} on both standard benchmarks and challenging in-the-wild sequences.
For quantitative evaluation, we utilize the \textbf{DexYCB} dataset~\cite{chao2021dexycb}, selecting 20 diverse trajectories that cover a wide range of object categories and interaction types.
To assess robustness under severe occlusion, we employ the \textbf{HO3D-v3} dataset~\cite{hampali2020honnotate}, testing on 18 sequences following the protocol of HOLD~\cite{fan2024hold}.
\revision{To evaluate generalization to bimanual interactions, we further include the \textbf{ARCTIC} dataset~\cite{fan2023arctic}, adopting its rigid-object subset following the HOLD protocol.}
Furthermore, to demonstrate generalization capabilities, we curate an \textbf{In-the-Wild} dataset.
This collection integrates sequences from prior work~\cite{fan2024hold} with self-captured footage, specifically targeting objects with complex geometries and intricate manipulation patterns to verify robustness in arbitrary settings.

\paragraph{Baselines.}
We benchmark our framework against state-of-the-art approaches in monocular hand-object reconstruction. 
Specifically, we compare with \textbf{HOLD}~\cite{fan2024hold}, which reconstructs interactions via implicit neural rendering, and \textbf{MagicHOI}~\cite{wang2025magichoi}, a recent method leveraging diffusion-based priors for geometry generation.
Detailed hyperparameters, implementation details, and runtime analysis are provided in the Supp.

\paragraph{Metrics.}
We evaluate system performance focusing on geometric fidelity, interaction plausibility, and robustness.
(1) \textit{Hand Accuracy}: We report the Root-relative Mean Per-Joint Position Error (MPJPE) in millimeters.
(2) \textit{Object Geometry}: Following the protocol of HOLD~\cite{fan2024hold}, we evaluate reconstruction quality using Chamfer Distance (CD) in squared centimeters ($cm^2$) and F-scores at 5mm and 10mm thresholds (F@5mm, F@10mm).
(3) \textit{Interaction Quality}: We compute the Hand-relative Chamfer Distance (CD$_h$), also in $cm^2$, by aligning the object to the hand's root frame to measure relative spatial consistency.
(4) \textit{Robustness}: Crucially, we report the \textit{Success Rate} (SR). A sequence is considered a failure if the method encounters initialization breakdown (e.g., SfM collapse) or suffers from catastrophic tracking drift.

\subsection{Comparisons}

\begin{figure*}[t] \centering
\includegraphics[width=\textwidth]{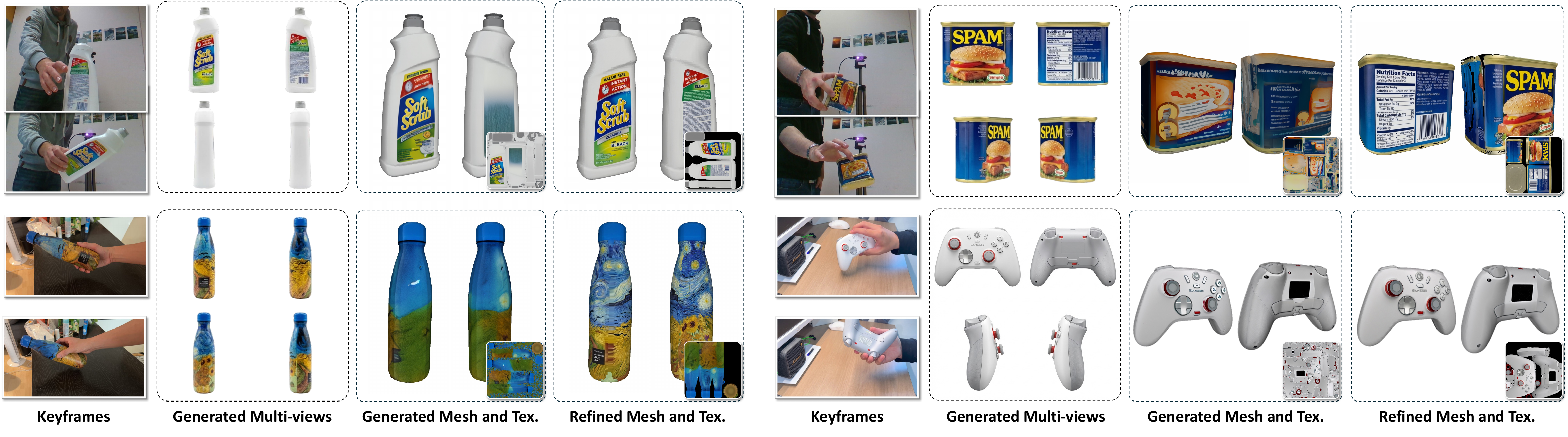}
    \caption{
\textbf{Qualitative Results of Agentic Generation.}
We visualize the intermediate stages of our pipeline across diverse object categories. Despite severe hand occlusion in the input keyframes, our VLM-guided approach successfully synthesizes consistent multi-view images and reconstructs high-fidelity 3D meshes. Notably, the \textit{texture refinement} step significantly enhances surface details and sharpness compared to the initial raw generation.
}
    \label{fig:gen_results}
\end{figure*}

\begin{figure*}[p] \centering
\includegraphics[width=0.96\textwidth,height=0.96\textheight,keepaspectratio]{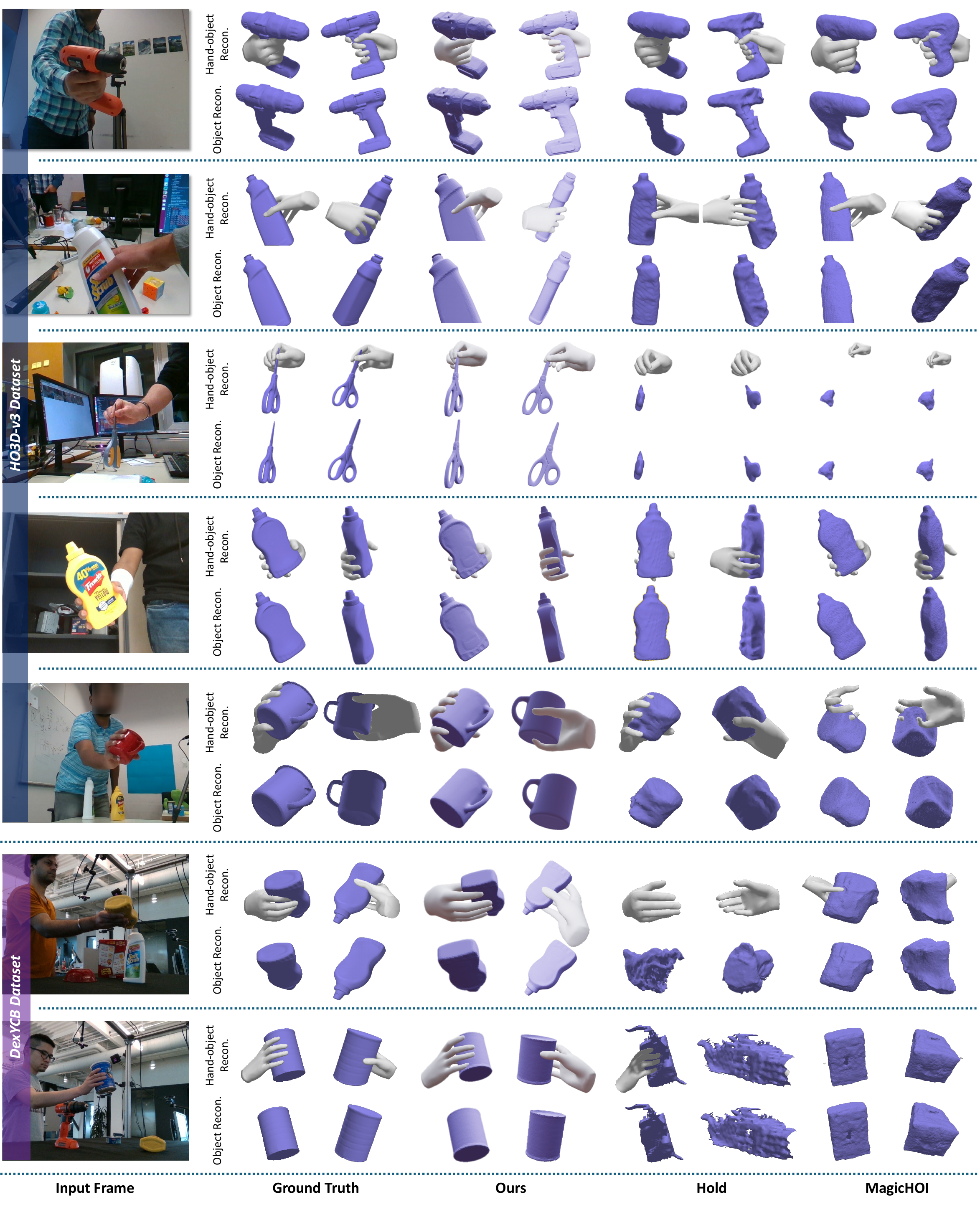}
    \caption{
    \textbf{Qualitative Comparison}. We compare our reconstructed hands and objects with baseline methods on the HO3D-v3 and DexYCB dataset, showing camera views as well as side views of the object-only and hand-object interaction results.
    }
    \label{fig:comp}
\end{figure*}

\begin{figure*}[t] \centering
\includegraphics[width=\textwidth]{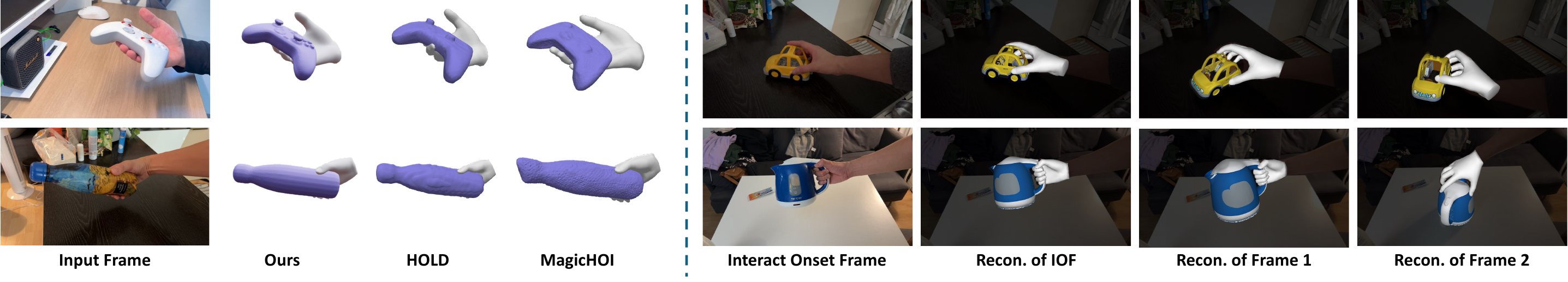}
    \caption{
    \textbf{Qualitative Evaluation on In-the-Wild Sequences.}
    \textit{(Left)} Comparison against state-of-the-art baselines. While HOLD~\cite{fan2024hold} and MagicHOI~\cite{wang2025magichoi} suffer from geometric noise or over-smoothed artifacts due to unreliable initialization, \sysname{} recovers clean, high-fidelity meshes.
    \textit{(Right)} Our reconstruction results across temporal sequences. Starting from the Interaction Onset Frame (IOF), our anchor-and-track strategy maintains robust alignment and physical plausibility throughout the dynamic interaction.
    }
    \label{fig:itw}
\end{figure*}

\begin{table*}[t]
    \centering
    \caption{
    \textbf{Quantitative comparison on DexYCB and HO3D-v3 datasets.}
    The \textbf{best} results are highlighted in bold, and the \underline{second best} are underlined.
    Baseline metrics ($\dagger$) are averaged over the \textit{successful subset} only, creating a survivor bias.
    }
    \label{tab:quant_comparison}
    \resizebox{1.0\textwidth}{!}{%
    \begin{tabular}{l ccccc c ccccc c}
        \toprule
        & \multicolumn{6}{c}{\textbf{DexYCB Dataset}} & \multicolumn{6}{c}{\textbf{HO3D-v3 Dataset}} \\
        \cmidrule(lr){2-7} \cmidrule(lr){8-13}
        Method & MPJPE (mm)$\downarrow$ & CD ($cm^2$)$\downarrow$ & F@5 (\%)$\uparrow$ & F@10 (\%)$\uparrow$ & CD$_h$ ($cm^2$)$\downarrow$ & SR (\%)$\uparrow$ 
               & MPJPE (mm)$\downarrow$ & CD ($cm^2$)$\downarrow$ & F@5 (\%)$\uparrow$ & F@10 (\%)$\uparrow$ & CD$_h$ ($cm^2$)$\downarrow$ & SR (\%)$\uparrow$ \\
        \midrule
        HOLD$^\dagger$~\cite{fan2024hold} 
        & 30.86 & 19.30 & 33.20 & 54.94 & \underline{170.9} & \underline{45.0} 
        & 22.09 & 1.11 & \underline{81.75} & \underline{92.42} & \underline{18.66} & \textbf{100.0} \\
        
        MagicHOI$^\dagger$~\cite{wang2025magichoi} 
        & \underline{21.20} & \underline{2.05} & \underline{45.67} & \underline{67.14} & 661.90 & 25.0 
        & \underline{7.38} & \underline{0.90} & 76.74 & 91.59 & 21.81 & 83.3 \\
        \midrule
        Ours 
        & \textbf{19.06} & \textbf{0.52} & \textbf{83.21} & \textbf{95.43} & \textbf{94.60} & \textbf{100.0} 
        & \textbf{3.92} & \textbf{0.27} & \textbf{86.63} & \textbf{97.77} & \textbf{15.81} & \textbf{100.0} \\
        \bottomrule
    \end{tabular}
    }
\end{table*}

\paragraph{Quantitative Evaluation.}
As presented in Table~\ref{tab:quant_comparison}, \sysname{} establishes a new state-of-the-art benchmark for monocular HOI reconstruction. 
We consistently outperform both optimization-based (HOLD) and generative-based (MagicHOI) baselines across all metrics.
In terms of geometric fidelity, \sysname{} achieves a Chamfer Distance (CD) of \textbf{0.52 $cm^2$} on DexYCB, reducing the reconstruction error by nearly \textbf{75\%} compared to MagicHOI (2.05 $cm^2$). 
This substantial gap highlights the advantage of our agentic generation pipeline over standard SfM-based initialization, which often yields noisy or incomplete geometry.

\paragraph{Interaction Stability and Drift.}
A critical differentiator lies in the interaction quality ($CD_h$). 
Prior methods suffer from severe \textit{interaction drift} due to scale ambiguity and lack of physical constraints. MagicHOI, for instance, exhibits a drastically high $CD_h$ of 661.90 $cm^2$, indicating that the object frequently loses contact or drifts far from the hand. 
In contrast, \sysname{} maintains a significantly lower $CD_h$ of 94.60 $cm^2$. 
This confirms that our \textit{Contact-Aware Optimization} effectively anchors the object to the hand, ensuring the reconstruction remains physically cohesive and valid for simulation.

\paragraph{Robustness and Survivor Bias.}
As shown in the Success Rate (SR) column, baselines exhibit high failure rates on DexYCB: MagicHOI fails on 75\% of the sequences, and HOLD fails on 55\%. 
It is important to note that the metrics reported for these baselines are biased, as they are averaged only over the \textit{easiest} subset of sequences where initialization succeeded (Survivor Bias). 
Conversely, \sysname{} achieves a 100\% Success Rate, evaluating on the entire dataset including the most challenging occlusion cases. 
Remarkably, even while averaging over these harder samples, \sysname{} still achieves a lower global pose error (MPJPE: 19.06 mm) than MagicHOI (21.20 mm) computed on its easy subset. 
This demonstrates that \sysname{} is not only more robust but also fundamentally more accurate.

\paragraph{Generation Quality.}
Figure~\ref{fig:gen_results} visualizes intermediate outputs of our agentic generation pipeline across diverse object categories. Despite severe hand occlusion in the input keyframes, the VLM-guided synthesis yields consistent multi-view images, and the texture refinement step substantially enhances surface details over the raw generation.

\paragraph{Qualitative Evaluation.}
Qualitative comparisons in Figure~\ref{fig:comp} corroborate the quantitative findings.
Baselines often produce fragmented geometry or suffer from "floating object" artifacts where the object drifts away from the interaction zone. 
Leveraging generative priors, \sysname{} reconstructs watertight, high-fidelity meshes. Moreover, the temporal consistency of our results eliminates the jitter observed in prior arts, faithfully reflecting the input video.

\paragraph{\revision{Bimanual Interactions Evaluation.}}
\revision{\sysname{} seamlessly extends to bimanual settings by jointly optimizing both hands with independent contact constraints. 
As Table~\ref{tab:arctic} demonstrates, our method establishes a new state-of-the-art, consistently outperforming both HOLD and BIGS~\cite{on2025bigs} across hand pose accuracy, object geometry, and interaction quality metrics. 
This confirms \sysname{}'s inherent robustness and its capability to generalize to complex multi-hand manipulations without requiring architectural modifications.}

\begin{table}[t]
    \centering
    \caption{
    \revision{\textbf{Quantitative comparison on the ARCTIC dataset (rigid-object subset).} \textbf{Best} results are in bold, \underline{second-best} are underlined.}
    }
    \label{tab:arctic}
    \resizebox{\columnwidth}{!}{%
    \begin{tabular}{l cc cc cc}
        \toprule
        Method & MPJPE$_l$$\downarrow$ & MPJPE$_r$$\downarrow$ & CD$_o$$\downarrow$ & F@5$\uparrow$ & CD$_l$$\downarrow$ & CD$_r$$\downarrow$ \\
        \midrule
        HOLD~\cite{fan2024hold} & \underline{27.1} & \underline{24.7} & 2.07 & 37.1 & 105.9 & 123.5 \\
        BIGS~\cite{on2025bigs} & 34.1 & 36.1 & \underline{1.36} & \underline{56.4} & \underline{46.1} & \underline{31.3} \\
        \midrule
        \textbf{Ours} & \textbf{25.0} & \textbf{23.8} & \textbf{1.12} & \textbf{57.6} & \textbf{21.9} & \textbf{30.6} \\
        \bottomrule
    \end{tabular}
    }
\end{table}

\subsection{Ablation Study}

\begin{figure*}[t] \centering
\includegraphics[width=\textwidth]{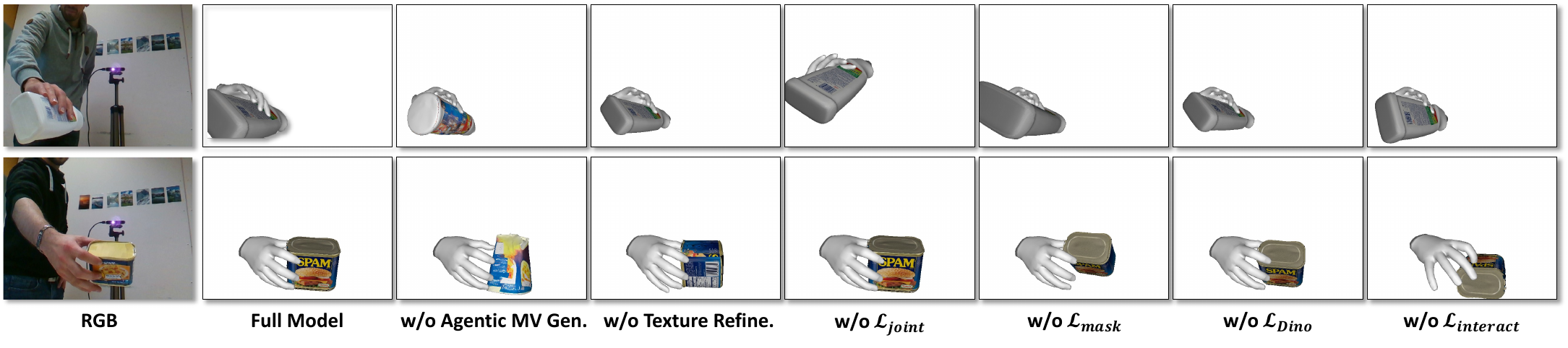}
    \caption{
\textbf{Qualitative Ablation Study.}
Visual comparisons demonstrate that removing key components---such as agentic generation or interaction constraints---leads to severe geometric artifacts, texture degradation, and physical violations (e.g., interpenetration), validating the necessity of our full pipeline.
}
    \label{fig:abl}
\end{figure*}

\begin{figure*}[t] \centering
\includegraphics[width=\textwidth]{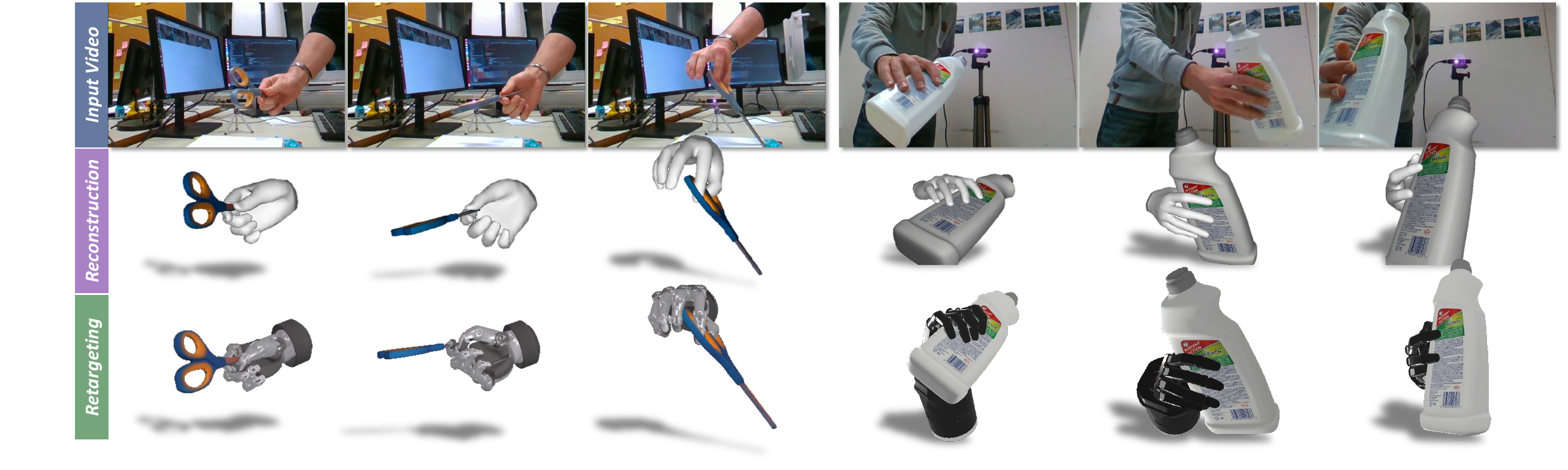}
    \caption{
    \textbf{Real-to-Sim retargeting results}, demonstrating that AGILE enables stable kinematic transfer of reconstructed human hand--object interactions to a multi-fingered robotic hand without physics-based correction.
    }
    \label{fig:retargeting}
\end{figure*}

\begin{table}[t]
    \centering
    \caption{Detailed ablation study on the HO3D dataset. We systematically evaluate the impact of generative components and optimization objectives.
    }
    \label{tab:ablation_study}
    \resizebox{1.0\columnwidth}{!}{
    \begin{tabular}{l l c cc c}
        \toprule
        & \multirow{2}{*}{Model Variant} & \multicolumn{1}{c}{Hand} & \multicolumn{2}{c}{Object Geometry} & \multicolumn{1}{c}{Interaction} \\
        \cmidrule(lr){3-3} \cmidrule(lr){4-5} \cmidrule(lr){6-6}
        & & MPJPE $\downarrow$ & CD $\downarrow$ & F@5mm $\uparrow$ & CD$_h$ $\downarrow$ \\
        \midrule
        (a) & \sysname{} (Full) & \textbf{3.92} & \textbf{0.27} & \textbf{86.63} & \textbf{15.81} \\
        \midrule
        \multicolumn{6}{l}{\textit{Generative Components}} \\
        (b) & w/o Agentic Multi-view Gen. & 4.02 & 2.85 & 30.12 & 40.12 \\
        \revision{(b$_1$)} & \revision{Heuristic Selection} & \revision{4.17} & \revision{2.04} & \revision{76.51} & \revision{29.37} \\
        \revision{(b$_2$)} & \revision{w/o Multi-View Synthesis} & \revision{4.32} & \revision{2.61} & \revision{66.58} & \revision{46.94} \\
        (c) & w/o Texture Refine. & 4.16 & 0.30 & 86.02 & 20.18 \\
        \midrule
        \multicolumn{6}{l}{\textit{Optimization Objectives}} \\
        (d) & w/o $\mathcal{L}_{\text{joint}}$ & 5.06 & 0.33 & 84.93 & 19.37 \\
        (e) & w/o $\mathcal{L}_{\text{mask}}$ & 3.97 & 0.40 & 81.14 & 23.99 \\
        (f) & w/o $\mathcal{L}_{\text{dino}}$ & 4.13 & 0.34 & 84.36 & 21.64 \\
        (g) & w/o $\mathcal{L}_{\text{interact}}$ & 4.17 & 0.35 & 86.10 & 54.40 \\
        \bottomrule
    \end{tabular}}
\end{table}

We conduct a detailed ablation study on the HO3D dataset to dissect the contribution of each component in our framework. 
We evaluate the impact on three key dimensions: Hand Pose Accuracy (MPJPE), Object Geometric Fidelity (CD), and Interaction Alignment (CD$_h$). 
The quantitative results are summarized in Table~\ref{tab:ablation_study}.

\textbf{(a) Full Model.} 
Our complete pipeline achieves the best performance across all metrics, validating the synergy between high-fidelity asset generation and physics-aware tracking.

\paragraph{Impact of Agentic Generation Modules.}
We first analyze the components responsible for creating the 3D asset, which serves as the foundation for subsequent tracking.

\noindent \textit{(b) w/o Agentic Multi-view Generation.}
\revision{Disabling the entire agentic loop (b) and lifting a 3D mesh directly from a single frame causes catastrophic geometric hallucinations in occluded regions. 
To further disentangle our agentic design from the innate capabilities of the base models, we ablate its two core components. 
Replacing the VLM critic with a max-mask heuristic (b$_1$) severely degrades accuracy, confirming that large masks often capture uninformative or self-occluded views rather than optimal geometric ones. 
Alternatively, retaining VLM selection but bypassing multi-view synthesis (b$_2$) similarly fails, as single-view generation forces the model to hallucinate unseen topologies. 
These substantial degradations prove that our performance gains are driven by the agentic paradigm, not merely the underlying foundation models.}

\noindent \textit{(c) w/o Texture Refine.}
Here, we skip the texture baking and refinement step. 
This ablation proves critical: high-quality texture is not merely for visual aesthetics but is a prerequisite for robust tracking. 
Since both our initialization (via FoundationPose) and our semantic consistency loss rely on discriminative surface features, noisy or blurred textures lead to initialization drift, causing a noticeable rise in interaction error (CD$_h$).

\paragraph{Impact of Optimization Objectives.}
Next, we evaluate the loss functions used during the decoupled optimization stages.

\noindent \textit{(d) w/o $\mathcal{L}_{\text{joint}}$.}
In this variant, the hand pose relies solely on the initial PnP estimation without further optimization. 
The significant degradation in MPJPE confirms that PnP initialization alone is insufficient for high-precision tracking. 
$\mathcal{L}_{\text{joint}}$ is indispensable for refining these initial estimates to resolve depth ambiguities and ensure temporally consistent hand articulation.

\noindent \textit{(e) w/o $\mathcal{L}_{\text{mask}}$.}
This component provides the strongest pixel-level geometric constraint for the rigid object. 
Although the object scale is pre-initialized, $\mathcal{L}_{\text{mask}}$ is critical for fine-tuning the scale at the anchor frame and guiding precise 6D pose optimization throughout the sequence. 
Without this constraint, the object pose fails to strictly adhere to visual boundaries, resulting in misalignment that negatively impacts both CD and F-scores.

\noindent \textit{(f) w/o $\mathcal{L}_{\text{dino}}$.}
We ablate the DINO-based semantic feature loss.
In the absence of this consistency term, object tracking becomes prone to drift, particularly in textureless regions or under motion blur. 
The increase in CD$_h$ confirms that semantic descriptors provide robust correspondence where pixel-level metrics alone are insufficient.

\noindent \textit{(g) w/o $\mathcal{L}_{\text{interact}}$.}
Finally, removing the interaction stability loss leads to severe physical violations. 
Without this constraint, the object tends to slide unnaturally or penetrate the hand mesh to minimize surface distance. 
Consequently, the fine-grained interaction quality degrades drastically (CD$_h$ spikes to 54.40), highlighting the necessity of SDF-based priors for physically plausible reconstruction.

\subsection{Application: Real-to-Sim Retargeting}
\label{sec:application}

To showcase the utility of our reconstructions for Embodied AI, we demonstrate a scalable Real-to-Sim retargeting pipeline.
We import the reconstructed sequence into Isaac Gym~\cite{makoviychuk2021isaac} and map the human hand motion to a dexterous robotic hand (e.g., Shadow) via kinematic optimization~\cite{qin2023anyteleop}, using the agentic-generated mesh as the manipulated asset.

As illustrated in Figure~\ref{fig:retargeting}, \sysname{} effectively bridges the gap between video and simulation:
(1) \textit{Simulation-Ready Assets}: Our generated meshes are watertight and topologically clean, ready to be loaded into the simulator without manual cleanup.
(2) \textit{Faithful Kinematic Transfer}: The retargeted grasps closely follow the recovered hand-object configurations through purely kinematic mapping, providing a qualitative validation that \sysname{} recovers accurate relative poses and contact states.
We note that this demonstration focuses on kinematic visualization rather than full physics-based rollout; closing the loop with contact dynamics and policy learning is left to future work. Nevertheless, by automating the path from monocular video to retargetable assets, our framework opens a route toward curating large-scale manipulation data from unconstrained videos for generalist policy learning.
\section{Conclusion}
\label{sec:conclusion}

\revision{In this work, we presented \sysname{}, a robust framework that addresses the persistent challenges of incomplete geometry and brittle initialization in monocular hand-object interaction reconstruction. By shifting the paradigm from \textit{reconstruction-based} optimization to \textit{agentic generation}, we leverage VLM-guided priors to synthesize high-fidelity, watertight object meshes even under severe hand occlusion. Complementing this, our \textit{anchor-and-track} strategy eliminates the dependency on fragile SfM pipelines, enabling stable 6D pose estimation on challenging in-the-wild videos where prior methods routinely fail. Extensive experiments on DexYCB, HO3D-v3, ARCTIC, and in-the-wild sequences confirm that \sysname{} achieves state-of-the-art accuracy with a 100\% success rate, producing simulation-ready digital twins that effectively bridge the gap between visual observation and physical simulation.

\paragraph{Limitations and Future Work.}
Our framework inherits limitations from the underlying foundation models for depth, segmentation, and multi-view generation, so extreme cases (e.g., transparent or highly reflective objects) may induce scale drift or local tracking errors. Notably, the static camera assumption applies only to the initial IOF detection and does not constrain core tracking, which remains valid under camera motion (see Supp.). Promising directions include extending agentic generation beyond rigid objects to articulated and deformable categories, and exploiting VLMs more broadly, both as a \textit{reconstructibility pre-filter} screening viable segments and as a \textit{multi-anchor detector} enabling bi-directional propagation against long-sequence drift.
}

\begin{acks}
This work was supported by the National Natural Science Foundation of China (No.~62576315).
\end{acks}

\bibliographystyle{ACM-Reference-Format}
\bibliography{main}

\clearpage
\appendix
\newcommand{\listingsubsubsection}[1]{%
\subsubsection{#1}\leavevmode\par\nobreak
}

\section{Implementation Details}
\label{sec:impl}
\paragraph{Pipeline and dependencies.}
Our framework integrates several state-of-the-art foundation models.
Crucially, we utilize \textbf{Gemini 3 Pro}~\cite{comanici2025gemini} as the VLM agent responsible for keyframe selection and rigorous quality assessment.
Guided by this agent, multi-view image synthesis is performed using the Gemini 2.5 Flash image generation model~\cite{comanici2025gemini}, while subsequent 3D mesh and texture generation are handled by Hunyuan 3D~\cite{hunyuan3d22025tencent, lai2025hunyuan3d25highfidelity3d}.
For preprocessing, geometric priors including metric depth and segmentation masks are extracted via MoGe-2~\cite{wang2025moge} and SAM2~\cite{ravi2024sam}, respectively, with the initial object pose estimated by FoundationPose~\cite{wen2024foundationpose}.
All differentiable rendering is implemented using PyTorch3D~\cite{ravi2020accelerating}.
The entire pipeline, from preprocessing to optimization, is designed to run efficiently on a single NVIDIA 4090 GPU.

\revision{
\paragraph{Automated retopology and UV unwrapping.}
After the initial 3D mesh is generated by Hunyuan 3D, we apply automated retopology and UV unwrapping to produce a clean, simulation-ready asset.
Specifically, we leverage Blender's built-in quad-remeshing algorithm to convert the raw triangular mesh into a uniform quad-dominant topology with a target face count of approximately 10,000 faces.
This step ensures the mesh has a regular, artifact-free topology suitable for physics simulation and collision detection.
Subsequently, we perform automatic UV projection using Blender's Smart UV Project, which segments the surface into non-overlapping UV islands and minimizes distortion.
The original texture is then baked onto the new UV layout via ray-cast projection, preserving the high-frequency surface details from the generative model.
This fully automated process eliminates the need for any manual post-processing.
}

\paragraph{Online optimization strategy.} Our method optimizes each frame sequentially via a two-stage alternating approach to ensure stability. Stage one optimizes hand translation (up to 200 iterations), followed by object 6D pose refinement (up to 400 iterations) with fixed hand parameters. exclusively for the onset frame, we optimize object scale by minimizing
$\mathcal{L}_{\text{mask}}$ and the ray-traced contact loss $\mathcal{L}_{\text{contact}}$.During scale optimization, object rotation is fixed, and translation is constrained to the visual ray passing through the object center, thereby distinctively regularizing the solution space. Convergence is reached when the moving average of parameter variations, including scale when active, is less than $10^{-4}$.

\paragraph{Optimizers and learning-rate schedule.} For optimization, we employ separate Adam optimizers~\cite{kingma2014adam} for rotation and translation parameters. The object optimization uses fixed learning rates of $2 \times 10^{-3}$ for rotation and $1 \times 10^{-3}$ for translation. Similarly, hand optimization uses fixed learning rates of $1 \times 10^{-3}$ for translation. We adopt constant learning rates throughout optimization, as we empirically observe that learning-rate decay does not lead to measurable accuracy improvements in our setting, while increasing the number of iterations required for convergence.

\paragraph{Loss terms and weights.} The loss function employs the following weights: $\mathcal{L}_{\text{mask}}$ (L2) is weighted at 5.0, $\mathcal{L}_{\text{dino}}$ (L1) at 10.0. For physical interaction modeling, the ray-traced contact loss $\mathcal{L}_{\text{contact}}$ receives a weight of 5.0. Additionally, the relative position interaction loss $\mathcal{L}_{\text{interact}}$ is weighted at $400$ for HO3D dataset and $200$ for DexYCB dataset, with a maximum distance threshold of 0.05m to prevent spurious long-range attractions between hand and object.
For the SDF-based distance-weight gating function $w_i = 1 - \tanh(\sigma \cdot d_i)$ defined in the main paper, we set $\sigma=40$.

\paragraph{Interaction Onset Frame selection.} For HO3D, we identify the interaction onset as the earliest frame where the object exhibits significant motion while remaining fully visible. We quantify the motion by measuring the variation of the non-occluded object mask between consecutive frames:
\begin{equation}
r_i=\frac{\lVert (\mathcal{M}_o^{i+1}-\mathcal{M}_o^{i})\odot\bigl(1-\mathcal{M}_h^{i+1}\bigr)\odot(1-\mathcal{M}_h^{i})\rVert_{1}}{\lVert \mathcal{M}_o^{i}\rVert_{1}+\varepsilon},
\end{equation}
where $\mathcal{M}_o$ and $\mathcal{M}_h$ denote the object and hand masks, respectively. The numerator captures the change in visible object pixels, while the denominator normalizes by the object area ($\varepsilon=10^{-8}$ ensures stability). We select the first frame $i$ that satisfies two criteria: (1) the motion ratio $r_i$ exceeds a threshold $\tau=0.025$, and (2) the object mask does not touch the image boundaries. The latter ensures the object is not truncated, thereby guaranteeing reliable geometric supervision.

\section{Why Start at Interaction Onset?}
\label{sec:start_frame_reason}

We choose the interaction onset frame as the optimization anchor for two key reasons. First, metric scale alignment: since monocular object reconstruction suffers from scale ambiguity, the physical contact allows us to leverage the hand's reliable metric scale to constrain and propagate the correct object size. Second, loss efficacy: the contact-based losses ($\mathcal{L}_{\text{contact}}$) are physically meaningful only when interaction occurs. Initiating optimization at this frame ensures these geometric constraints provide valid gradients for precise hand--object alignment.

\revision{
\section{Clarification on the Static Camera Assumption}
\label{sec:static_camera}

We clarify that the static camera assumption in our framework is strictly limited to the automatic selection of the Interaction Onset Frame (IOF), rather than the core reconstruction pipeline itself.

\paragraph{Why a static camera for IOF detection?}
As described in Sec.~\ref{sec:impl}, we identify the IOF by measuring the 2D pixel displacement of the unoccluded object mask between consecutive frames. This specific heuristic assumes a fixed camera to reliably distinguish genuine object movement from camera ego-motion. Under a moving camera, apparent mask displacement could arise from viewpoint change rather than object manipulation, leading to incorrect IOF identification.

\paragraph{Core pipeline independence.}
Crucially, our core pipeline---Agentic Generation (Sec.~3.1), Pose Initialization (Sec.~3.2), and Contact-Aware Tracking (Sec.~3.3)---operates entirely in the camera coordinate system. Since all optimization variables (hand pose, object pose, and interaction constraints) are defined relative to the camera frame, camera ego-motion is mathematically irrelevant to our optimization. This means the reconstruction quality is not affected by whether the camera is static or dynamic, as long as a valid IOF is identified.

\paragraph{Extension to dynamic cameras.}
To handle dynamic cameras (e.g., first-person or egocentric videos), one simply needs to substitute the IOF selection step with alternative methods---such as VLM-based action recognition (e.g., detecting the moment of grasping), off-the-shelf contact predictors, or manual specification. Camera intrinsics and extrinsics for each frame can be recovered using existing structure-from-motion or SLAM pipelines (e.g., MegaSaM~\cite{li2025megasam}). The rest of our framework requires absolutely zero modification. We view this modularity as a strength: the IOF selection is a replaceable preprocessing component, while the core method is camera-motion agnostic.
}

\revision{
\section{Details of Fine-Grained Ablation Variants}
\label{sec:ablation_agentic}

The fine-grained ablation results on Agentic Multi-view Generation are reported in Table~3 of the main text (variants b$_1$, b$_2$). Here we describe the ablation configurations in detail:

(1) \textit{Heuristic Selection} (b$_1$): We replace the VLM-based keyframe selection with a heuristic that simply chooses the frames with the largest object mask areas. Multi-view synthesis and all subsequent steps remain unchanged. This isolates the contribution of the VLM's spatial reasoning for selecting geometrically informative views.

(2) \textit{w/o Multi-View Synthesis} (b$_2$): We retain VLM keyframe selection but skip the multi-view synthesis stage, directly feeding the single selected keyframe to the 3D generation model. This isolates the contribution of multi-view consistency for faithful 3D reconstruction.

Both variants use the \emph{exact same} foundation models (Gemini, Hunyuan3D) as the full pipeline to ensure a fair comparison.
}

\section{Sequence Used}
\label{sec:seq}

\begin{table}[t]
    \centering
    \caption{Sequences selected from the HO3D datasets.}
    \label{tab:ho3d_sequences}
    \resizebox{0.7\columnwidth}{!}{
    \begin{tabular}{llc}
        \toprule
        Sequence ID & Object & IOF Index \\
        \midrule
        ABF12 & bleach cleanser & 100 \\
        ABF14 & bleach cleanser & 100 \\
        BB12 & banana & 200 \\
        BB13 & banana & 130 \\
        GPMF12 & potted meat & 105 \\
        GPMF14 & potted meat & 100 \\
        GSF13 & scissors & 100 \\
        GSF14 & scissors & 80 \\
        MC1 & cracker box & 0 \\
        MC4 & cracker box & 0 \\
        MDF12 & power drill & 150  \\
        MDF14 & power drill & 250 \\
        ShSu10 & sugar box & 50 \\
        ShSu12 & sugar box & 110 \\
        SM2 & mustard & 0 \\
        SM4 & mustard & 0 \\
        SMu1 & mug & 20 \\
        SMu40 & mug & 0 \\
        \bottomrule
    \end{tabular}}
\end{table}
        
\begin{table}[h]
    \centering
    \caption{Sequences selected from the DexYCB datasets.}
    \label{tab:dexycb_sequences}
    \resizebox{\columnwidth}{!}{
    \begin{tabular}{llc}
        \toprule
        Sequence ID & Object \\
        \midrule
        20200709-subject-01/20200709\_141754/836212060125 & master chef can \\
        20200709-subject-01/20200709\_142853/836212060125 & master chef can \\
        20200813-subject-02/20200813\_154204/836212060125 & bleach cleanser \\
        20200813-subject-02/20200813\_153453/836212060125 & power drill\\
        20200820-subject-03/20200820\_141856/836212060125 & potted meat \\
        20200820-subject-03/20200820\_143330/836212060125 & mug \\
        20200903-subject-04/20200903\_103554/836212060125 & master chef can \\
        20200903-subject-04/20200903\_113012/836212060125 & wood block \\
        20200908-subject-05/20200908\_144138/836212060125 & tomato soup can \\
        20200908-subject-05/20200908\_152416/836212060125 & large clamp \\
        20200918-subject-06/20200918\_113441/836212060125 & sugar box \\
        20200918-subject-06/20200918\_114747/836212060125 & potted meat \\
        20200928-subject-07/20200928\_153800/836212060125 & pitcher base \\
        20200928-subject-07/20200928\_144300/836212060125 & mustard bottle \\
        20201002-subject-08/20201002\_110940/836212060125 & pitcher base \\
        20201002-subject-08/20201002\_105558/836212060125 & mustard bottle \\
        20201015-subject-09/20201015\_143700/836212060125 & mustard bottle \\
        20201015-subject-09/20201015\_142940/836212060125 & cracker box \\
        20201022-subject-10/20201022\_111745/836212060125 & mustard bottle \\
        20201022-subject-10/20201022\_111209/836212060125 & sugar box \\
        \bottomrule
    \end{tabular}}
\end{table}

\begin{table}[t]
\centering
\caption{\revision{Comparison of computational cost for a 200-frame sequence on a single RTX 4090 GPU. Our method requires significantly less preprocessing and total runtime than existing methods.}}
\label{tab:runtime}
\begin{tabular}{lccc}
\toprule
Method & Preprocessing & Optimization & Total \\
\midrule
HOLD & $\sim$65 min & $\sim$24 h & $\sim$25.1 h \\
MagicHOI & $\sim$25 min & $\sim$2 h & $\sim$2.4 h \\
Ours & \revision{$\sim$21--29 min} & \revision{$\sim$1.75 h} & \revision{$\sim$\textbf{2.1--2.2 h}} \\
\bottomrule
\end{tabular}
\end{table}

We evaluate our method on sequences from the DexYCB~\cite{chao2021dexycb} and HO3D~\cite{hampali2020honnotate} datasets.
Specifically, as shown in Table~\ref{tab:ho3d_sequences} and Table~\ref{tab:dexycb_sequences}, we randomly select 18 sequences from HO3D and 20 sequences from DexYCB, covering a diverse range of object types and hand-object interaction patterns.
For the HO3D dataset, we begin processing from the identified IOF to ensure meaningful interaction modeling. HO3D sequences vary in length from 800 to 1600 frames; for computational efficiency, we cap the processing at 1024 frames per sequence. For the DexYCB dataset, each selected sequence spans approximately 80 frames. Unless otherwise specified, we uniformly subsample all sequences with a stride of 5 (i.e., processing every fifth frame), which balances temporal coverage with computational cost.
These sequences are chosen to provide comprehensive coverage of common manipulation scenarios and object categories, enabling robust evaluation of our method's generalization capabilities across different interaction types and geometric configurations.

\section{Analysis on Results of Baselines}
\label{sec:baselines_analysis}
On the DexYCB dataset, both baselines exhibited varying degrees of failure. Specifically, HOLD encountered issues in some sequences where it failed to obtain the hand/object mesh due to inaccurate poses, which prevented the geometric structure from being effectively learned. Meanwhile, MagicHOI was unable to complete colmap reconstruction in most sequences and thus could not obtain stable poses for subsequent processing. On the other hand, MagicHOI's hand-object alignment relies on the geometry generated by its generative model. Due to the limitations of its capabilities, it fails to produce reasonable geometric structures in some challenging cases, leading to the failure of the final results.

\section{Computation Cost}
\label{sec:cost}
Our optimization process is computationally efficient. On a single NVIDIA RTX 4090 GPU, each frame requires approximately 30--50 seconds for optimization. Unless otherwise specified, we process every fifth frame of each sequence. As a result, the total computation time scales linearly with the sequence length. On average, a sequence of 1000 frames (200 frames after downsampling) can be processed in approximately 1.5 hours, which is consistent with the overall runtime reported in Table~\ref{tab:runtime}.

\revision{
\paragraph{Detailed runtime breakdown.}
We report the end-to-end runtime for a representative 200-frame sequence (1000 raw frames at stride 5) on a single NVIDIA RTX 4090 GPU. Despite introducing a VLM critic in the generation loop, our total pipeline time remains competitive with or faster than existing baselines:

\begin{itemize}[leftmargin=1.5em, itemsep=2pt]
    \item \textbf{HOLD}: $\sim$25.1 hours (Preprocessing: $\sim$65 min; Optimization: $\sim$24 h)
    \item \textbf{MagicHOI}: $\sim$2.4 hours (Preprocessing: $\sim$25 min; Optimization: $\sim$2 h)
    \item \textbf{Ours}: $\sim$\textbf{2.1--2.2 hours} (Preprocessing: $\sim$21--29 min; Optimization: $\sim$1.75 h)
\end{itemize}

\textit{Breakdown of our preprocessing ($\sim$21--29 min):}
The Agentic Generation stage accounts for $\sim$16--24 minutes, depending on the number of rejection sampling iterations ($n \in [1, 3]$). This includes keyframe selection ($\sim$2 min), 3D generation ($\sim$10 min), multi-view synthesis ($\sim$2--6 min), and texture refinement ($\sim$2--6 min). The remaining preprocessing takes $\sim$5 minutes: mask extraction via SAM2 ($\sim$20 s), metric depth via MoGe-2 ($\sim$88 s), hand pose via WiLoR ($\sim$24 s), and FoundationPose initialization ($\sim$160 s).
}

\section{Details of VLM-Guided 3D Generation}
\label{sec:vlm_details}

In this section, we provide the detailed prompt specifications and evaluation protocols used in our agentic generation pipeline. Our framework leverages a Vision-Language Model (VLM) as an intelligent supervisor to guide three critical stages: (1) informative keyframe selection, (2) consistent multi-view synthesis, and (3) high-fidelity texture refinement. For each stage, we present the system instructions, task definitions, and representative VLM responses. The prompts presented below are the exact instructions used in our pipeline to ensure reproducibility.

\subsection{Stage 1: Informative Keyframe Selection}

The objective of this stage is to filter the input video stream and identify a minimal set of frames that maximize geometric coverage while minimizing occlusion.

\listingsubsubsection{Prompt Specification}

\begin{lstlisting}[language=json, caption={Prompt for Keyframe Selection}]
{
  "system_role": "You are an expert in computer vision, specializing in selecting frames that best showcase the complete appearance of an object from multiple images.",
  "task": "Select {num_frames} images from the following candidate frames. The selected images must:
  1. Clearly display different angles of the object.
  2. Maximize coverage of the object's complete appearance (front, back, left, right, top, bottom).
  3. Be sharp, with the object fully visible and minimal occlusion.
  4. Have the maximum possible viewpoint difference between selected frames.
  5. Feature the object occupying a relatively large portion of the frame.",
  "response_format": "
  {
    "selected_frames": [1, 5, 10, 15],  // Indices of selected frames
    "reasoning": "Reason for selection",
    "coverage": {
      "front": true,
      "back": true,
      "left": true,
      "right": false
    }
  }",
  "instruction": "Please reply strictly in JSON format without adding other explanatory text."
}
\end{lstlisting}

\listingsubsubsection{Example Response}

\begin{lstlisting}[language=json, caption={VLM Response for Keyframe Selection}]
{
  "selected_frames": [1, 5, 13, 21],
  "reasoning": "Frame 1 is selected as the standard front view, clearly displaying the main brand label. Frame 5 shows the back of the object, clearly presenting the ingredients list and nutritional information. Frame 13 provides the side profile, effectively showcasing the object's depth, thickness, and side curves. Frame 21 captures the bottom perspective, supplementing this often-overlooked angle. All four images are high-definition, have a large object-to-frame ratio, and exhibit significant viewpoint differences, achieving maximum multi-angle coverage.",
  "coverage": {
    "front": true,
    "back": true,
    "left": true,
    "right": false,
    "top": false,
    "bottom": true
  }
}
\end{lstlisting}

\subsection{Stage 2: Multi-View Synthesis and Validation}

Based on the selected keyframes, we generate four orthogonal views. To ensure consistency, a VLM critic evaluates the generated views against strict geometric and texture criteria.

\listingsubsubsection{Generation Prompts}

\begin{lstlisting}[language=json, caption={Prompts for Multi-View Generation}]
{
  "multiview_generation": {
    "static_object": "Generate a complete image of the water bottle in the picture from four perspectives: front, back, left, and right. Require photorealism. Present as a four-view grid. Remove the plate and extraneous objects. Pure white background. The image must not contain text indicating viewpoint direction (e.g., front, back, left, right).",
    "hand_held_object": "Generate a complete image of the object held in the hand from four perspectives: front, back, left, and right. Require photorealism. Present as a four-view grid. Remove the hand and hallucinate/complete the regions occluded by the hand. Pure white background. The image must not contain text indicating viewpoint direction. The object size should be consistent across all four views."
  }
}
\end{lstlisting}

\listingsubsubsection{Validation Protocol (The "Critic")}

\begin{lstlisting}[language=json, caption={Prompt for Multi-View Validation}]
{
  "validation": {
    "system_role": "You are an expert in 3D modeling, material analysis, and multi-view image quality assessment. Please carefully analyze the following images:",
    "task_description": "Task:
    1. The first image(s) are the original input, showing the appearance, texture, and material of an object.
    2. The last image is the generated 'four-view' image, which should display the complete image of the object from four different perspectives (front, back, left, right) while preserving the original visual attributes.",
    "evaluation_criteria": "Criteria:

    Level 1: Veto Items
    1. Text Check: Does the generated image contain any text, labels, or viewpoint descriptions (e.g., front, back)? If yes, terminate evaluation; result is invalid.

    Level 2: Core Dimension Scoring (0-10)
    2. Geometry & View Correctness (Weight: 30%): Are viewpoints correct? Is orientation consistent (no rotation)? Any rotation results in large deductions.
    3. Texture & Material Fidelity (Weight: 20%): Are surface textures (e.g., patterns) and material properties (e.g., reflection) consistent with the original?
    4. Geometric Detail Integrity (Weight: 20%): Are key geometric details (chamfers, holes, embossing) preserved?
    5. Feature Consistency (Weight: 15%): Is it the same object in terms of shape, style, and color?
    6. Image Quality (Weight: 15%): Is the image clear, noise-free, and on a pure white background?

    Level 3: Deductions
    - Rotated views: -3 points each.
    - Poor layout: -1 to -2 points.",
    "response_format": "JSON format containing: is_valid, score_overall, score_breakdown, has_text, rotated_views, improvement_suggestions, summary_feedback, etc.",
    "instruction": "Reply strictly in JSON. First check veto items."
  }
}
\end{lstlisting}

\subsection{Stage 3: Texture Refinement and Agentic Editing}

After lifting the multi-view images to 3D and unwrapping UVs, we perform a texture refinement step. A VLM critic detects artifacts (e.g., hallucinated content, missing details) and guides a subsequent editing pass.

\listingsubsubsection{Texture Validation Protocol}

\begin{lstlisting}[language=json, caption={Prompt for Texture Map Validation}]
{
  "texture_validate":{
    "system_role": "You are an expert in 3D texture mapping analysis, material comparison, and image consistency assessment. Carefully analyze the images.",
    "task_description": "Task:
    1. Image 1: The four-view image.
    2. Image 2: The generated Texture Map.
    3. Evaluate if the texture map accurately reproduces all texture information from the four-view image, focusing on completeness, fictional content, and correspondence.",
    "evaluation_criteria": "Criteria:

    Level 1: Veto Items
    - Invalid texture map (blank, pure color, severe distortion).
    - Key features completely missing.

    Level 2: Core Dimensions
    1. Completeness (Weight: 30%): Are all visible textures present? (Deduction: -2 per major missing item).
    2. Accuracy/No Fiction (Weight: 25%): Does it contain hallucinated content not present in the source? (Severe penalty: -3 to -5 points).
    3. Correspondence (Weight: 20%): Are textures mapped to correct UV islands?
    4. Color/Material (Weight: 15%): Consistency in color, shading, and saturation.
    5. Fidelity (Weight: 10%): Resolution and detail preservation.

    Level 3: Extra Deductions
    - Seams, repetitions, stretching.",
    "response_format": "JSON format containing: is_valid, score_overall, fictional_content_analysis, missing_content_analysis, texture_coverage_analysis, etc.",
    "instruction": "Strictly follow JSON format. Pay special attention to 'fictional content', any texture appearing in the map but not in the source views is a severe error!"
  }
}
\end{lstlisting}

\listingsubsubsection{Example Validation Result (Invalid Case Detected)}

\begin{lstlisting}[language=json, caption={Example Validation Result detecting hallucinations}]
{
  "is_valid": false,
  "score_overall": 0.0,
  "fictional_content_analysis": {
    "has_fictional_content": true,
    "fictional_items": [
      {
        "type": "severe",
        "description": "The entire back texture is a fabricated copy of the front label instead of the nutrition facts panel shown in the source.",
        "location_in_texture": "The second large body island in the texture map.",
        "deduction": 5
      }
    ],
    "total_fictional_deduction": 5
  },
  "summary_feedback": "Invalid. Critical error: the back side texture is a fabrication (copy of the front). This makes the model unusable for pose estimation from rear viewpoints."
}
\end{lstlisting}

\listingsubsubsection{Refinement Prompt Construction}

Based on the negative feedback from the validation step, we programmatically construct a structured refinement command to guide the image-to-image editing model.

\begin{lstlisting}[language=json, caption={Structured Input for Texture Refinement}]
{
  "texture_refinement_task": {
    "base_instruction": "Below are object images from multiple views and their texture map. Please edit the texture map based on the following critique:",
    "injected_feedback": "Invalid. Critical error: the back side texture is a fabrication (copy of the front). This makes the model unusable for pose estimation from rear viewpoints.",
    "geometric_constraints": "Do not change the layout or shape of UV islands. Only align the content with reference images. The bottom-left area corresponds to the bottom view.",
    "specific_corrections": "Change the purple/black artifact regions to match the red color of other parts. Adjust tone for vibrancy. Fix text and patterns."
  }
}
\end{lstlisting}

\end{document}